\documentclass[document]{ieeeconf}

\IEEEoverridecommandlockouts

\makeatletter
\let\NAT@parse\undefined
\makeatother

\usepackage{amssymb,amsmath}

\usepackage[sort&compress,numbers,sectionbib]{natbib}
\usepackage{cite} 
\usepackage{booktabs}
\usepackage{caption}
\usepackage{subcaption}
\usepackage{graphicx}

\usepackage{verbatimbox}
\usepackage{multirow}
\usepackage{balance}
\usepackage[table,xcdraw,dvipsnames]{xcolor}

\newcommand{\figref}[1]{Fig.~\ref{#1}}

\usepackage[hyphens]{url}

\usepackage{hyperref}
\hypersetup{bookmarksopen,bookmarksnumbered,
pdfpagemode=UseOutlines,
colorlinks=true,
linkcolor=teal,
anchorcolor=teal,
citecolor=teal,
filecolor=teal,
menucolor=teal,
urlcolor=teal,
breaklinks=true
}
\definecolor{myred}{RGB}{215,48,39}
\definecolor{myblue}{RGB}{69,117,180}
\definecolor{myorange}{RGB}{252,141,89}
\definecolor{mylightblue}{RGB}{145,191,219}

\usepackage{color,soul} 
\definecolor{krishna}{rgb}{.91,.65,.60} 
\definecolor{krishna-text}{rgb}{.59,.24,.19} 
\usepackage{svg}

\newcommand{\xxnote}[3]{}
\ifx\hidenotes\undefined
  \usepackage{color}
  \renewcommand{\xxnote}[3]{\color{#2}{#1: #3}}
\fi

\usepackage[normalem]{ulem}
\useunder{\uline}{\ul}{}
\usepackage{todonotes}

\usepackage{microtype}

\title{\LARGE \bf
From Crowd Motion Prediction to Robot Navigation in Crowds
}

\author{Sriyash Poddar$^{1}$, Christoforos Mavrogiannis$^{2}$, Siddhartha S. Srinivasa$^{2}$
\thanks{$^{1}$Department of Computer Science and Engineering, Indian Institute of Technology, Kharagpur, Kharagpur, India. Email: \texttt{poddarsriyash@iitkgp.ac.in}}
\thanks{$^{2}$Paul G. Allen School of Computer Science \& Engineering, University of Washington, Seattle, USA. Email: \texttt{\{cmavro, siddh\}@cs.washington.edu}}
\thanks{This work was (partially) funded by the Honda Research Institute USA, the National Science Foundation NRI (\#2132848) and CHS (\#2007011), DARPA RACER (\#HR0011-21-C-0171), the Office of Naval Research (\#N00014-17-1-2617-P00004 and \#2022-016-01 UW), and Amazon.}
}

\begin{document}

\maketitle
\thispagestyle{empty}
\pagestyle{empty}



\begin{abstract}

We focus on robot navigation in crowded environments. To navigate safely and efficiently within crowds, robots need models for crowd motion prediction. Building such models is hard due to the high dimensionality of multiagent domains and the challenge of collecting or simulating interaction-rich crowd-robot demonstrations. While there has been important progress on models for offline pedestrian motion forecasting, transferring their performance on real robots is nontrivial due to close interaction settings and novelty effects on users. In this paper, we investigate the utility of a recent state-of-the-art motion prediction model (S-GAN) for crowd navigation tasks. We incorporate this model into a model predictive controller (MPC) and deploy it on a self-balancing robot which we subject to a diverse range of crowd behaviors in the lab. We demonstrate that while S-GAN motion prediction accuracy transfers to the real world, its value is not reflected on navigation performance, measured with respect to safety and efficiency; in fact, the MPC performs indistinguishably even when using a simple constant-velocity prediction model, suggesting that substantial model improvements might be needed to yield significant gains for crowd navigation tasks. Footage from our experiments can be found at \url{https://youtu.be/mzFiXg8KsZ0}.

\end{abstract}


\section{Introduction}\label{sec:intro}

Large-scale deep learning architectures~\citep{CasasLU18, djuric2018motion, soo2016egocentric,rudenko2019-predSurvey,trajectron, gupta2018social, zhang19srlstm}
have been dramatically improving the state-of-the-art in prediction accuracy across standard benchmarks~\citep{PellegriniESG09,UCY}. While these models have been the foundation underlying recent real-world robot demonstrations~\citep{hoermann2017dynamic,chen2019crowd,liu2020decentralized,chen2020relational, Everett18_IROS, brito-ral21}, scaling their performance to complex environments like pedestrian domains, warehouses, or hospitals is challenging as these environments feature close interaction settings, a large space of behavior, and limited rules. 




\begin{figure}
\centering
    \includegraphics[width = \linewidth]{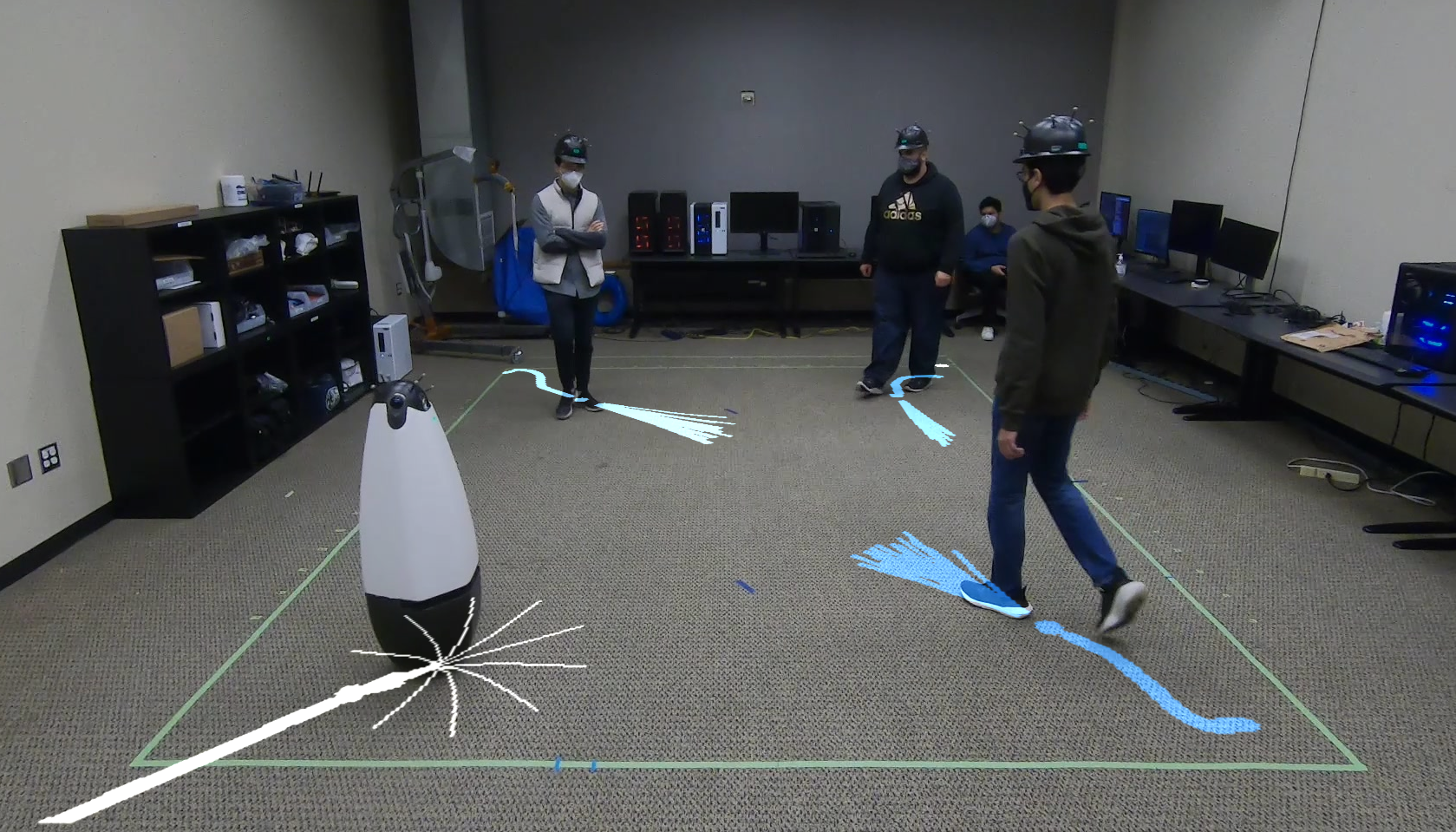}
    \caption{Honda's experimental ballbot~\citep{pathbot} navigates next to three users in our lab. Agents' past trajectories and distribution of future actions are shown. In this paper, we approach the question of how human motion prediction accuracy translates into robot navigation performance in crowded environments. \label{fig:introfig}}
\end{figure}



To address these challenges, many approaches involve training deep-learning models on simulated crowd-robot interactions~\citep{chen2019crowd,liu2020decentralized,chen2020relational, Everett18_IROS}. While typically employed crowd simulators~\citep{Helbing95,ORCA} produce realistically looking crowd behaviors, some of their core assumptions limit their relevance to crowd-navigation tasks. For instance, \citet{fraichard2020} showed that the assumptions of omniscience and homogeneity of existing crowd simulators give rise to behaviors that would be unsafe to execute on a real robot. Further, \citet{core-challenges2021} showed that a non-reactive, non-collision-avoiding agent is safer than ORCA-simulated agents in an ORCA-simulated world~\citep{ORCA} due to the overly submissive behaviors this model may exhibit. 

Other approaches use pedestrian datasets to train and validate models for crowd motion prediction \citep{SunM-RSS-21, Mangalam2021FromGW}. However, the pedestrian datasets most commonly used~\citep{PellegriniESG09,UCY} feature well-structured, goal-directed, and cooperative motion. These settings represent a very narrow subset of the behavior that a robot would encounter in the real world. This behavior is so prevalent in those datasets that according to \citet{scoller2019cv}, even constant-velocity (CV) prediction, a very simple, analytical model, performs comparably to recent state-of-the-art (SOTA) deep models. Therefore, while the SOTA in human motion prediction keeps improving, it is unclear what its relevance is for robot navigation in crowds.


Inspired by these observations, we ask the question: 

\begin{quote}\emph{To what extent does crowd motion prediction accuracy translate to robot navigation performance in crowd navigation tasks?}\end{quote}
To approach this question, we investigate the transfer of a recent SOTA model (S-GAN~\citep{gupta2018social}) from offline datasets to onboard performance and its implications on navigation performance. We integrate the S-GAN as a dynamics model into a standard MPC architecture and deploy it on a self-balancing robot, which we subject to a series of diverse crowd-robot interactions in the lab. We find that while the onboard prediction accuracy of S-GAN is superior to a simple CV baseline, the MPC navigation performance (measured in terms of safety and time efficiency) is indistinguishable, suggesting that substantial prediction model improvements may be needed to achieve improved navigation performance. 

\section{Related Work}\label{sec:relatedwork}


We discuss related work from the human motion prediction and crowd navigation literature.

\subsection{Crowd Motion Prediction}



The goal of modeling interactions among crowds in pedestrian domains has motivated much of the recent work in human motion prediction~\citep{rudenko2019-predSurvey}. Recent works have used a variety of architectures including recurrent~\citep{zhang19srlstm, Becker2018REDAS} and convolutional~\citep{Nikhil2018ConvolutionalNN} neural networks, spatiotemporal graphs~\citep{trajectron,roh2020corl,liu2021gcn}, state-refinement modules~\citep{sathyamoorthydensecavoid}, explicit probability maps~\citep{Mangalam2021FromGW}, normalizing flows~\citep{bhattacharya2019confitional}, and Gaussian processes~\citep{trautmanijrr, SunM-RSS-21}. Some of the recent state-of-the-art architectures are based on generative adversarial networks (GAN)~\citep{sophie19,gupta2018social}. These are particularly applicable to motion prediction tasks due to their ability to model the multimodality and diversity of crowd navigation domains. 

Inspired by the effectiveness of GAN-based approaches, we build our crowd motion prediction architecture around S-GANs~\citep{gupta2018social}. While prior work has used S-GANs primarily for motion tracking on offline datasets~\citep{gupta2018social} and simulated environments~\citep{wang2021corl}, in this work we deploy a S-GAN-based architecture on a real robot navigating under a variety of crowd conditions. Our implementation enables real-time performance capable of handling dynamic environments. 



\subsection{Crowd Navigation}

In recent years, several crowd navigation algorithms have been deployed on real robots~\citep{core-challenges2021}. Some approaches incorporate explicit models of human motion prediction into receding-horizon reactive controllers~\citep{trautmanijrr,ziebart2009,kretzschmar_ijrr16,chen2020relational,mavrogiannis2021topologyinformed,wang2021corl,brito-ral21,SunM-RSS-21}. Others learn end-to-end navigation policies using techniques like deep reinforcement learning~\citep{chen2019crowd, Everett18_IROS, liu2020decentralized,chen2020relational}

Our approach falls into the former category: similar to several recent works~\citep{brito-ral21,wang2021corl,mavrogiannis2021topologyinformed}, we integrate a crowd motion prediction model into a MPC architecture. In our prior work, we showed that CV-based motion prediction can empower a MPC to outperform recent end-to-end approaches~\citep{mavrogiannis2021topologyinformed}. In this work, we explore the utility of the recent state-of-the-art architectures like S-GANs for crowd navigation tasks.





\subsection{Benchmarking in Crowd Navigation}

One challenge in crowd navigation research is benchmarking and validation~\citep{core-challenges2021}. Observing the limitations of widely adopted practices as reported in recent literature~\citep{fraichard2020,core-challenges2021,scoller2019cv}, some works have developed new simulation environments~\citep{socnavbench,Grzeskowiak,Tsoi_2021_Sean_EP}, real-world datasets~\citep{tbd-dataset}, and experimental protocols~\citep{social-momentum-thri,mavrogiannis2021topologyinformed,pirk2022protocol} to improve the validation of future frameworks.

In this work, we also contribute towards these efforts by developing a series of benchmarking experiments designed to subject a navigation system to diverse crowd conditions. Unlike prior work, which typically focuses on navigation under cooperative, goal-directed settings, in this paper, we also develop benchmarking scenarios in non-cooperative settings, where humans are \emph{aggressive} or \emph{distracted} during navigation.

\section{Problem Statement}\label{sec:preliminaries}

We consider a workspace $\mathcal{W}\subseteq\mathbb{R}^2$ where a robot navigates among $n$ human agents. We denote by $s\in\mathcal{W}$ the state of the robot and by $s^{i}\in\mathcal{W}$ the state of agent $i\in\mathcal{N}=\{1, \dots n\}$. The robot is navigating from a state $s_{0}$ towards a goal state $g$ whereas agent $i\in\mathcal{N}=\{1, \dots n\}$ is navigating from $s_{0}^i$ towards a destination $g^i$. The robot is not aware of agents' destinations but we assume that it is fully observing the complete world state $(s_t, s^{1:n}_t)$ at every timestep $t$. By maintaining a history of states for all agents, the robot predicts their future trajectories using a model $f$. In this paper, our goal is to investigate whether the prediction accuracy of $f$ translates to robot navigation performance. As a proxy for navigation performance, we consider metrics capturing safety and efficiency properties of robot motion.

\section{Human Motion Prediction}\label{sec:prediction}


We treat human motion prediction as trajectory prediction over a finite horizon $T$ given the observation of a partial trajectory of horizon $h$.





\subsection{Probabilistic Modeling}\label{sec:probabilisticmodel}


We denote by $s^i_{t-h:t}\in\mathcal{W}^{h}$ the partial trajectory of an agent $i\in\mathcal{N}$ of horizon $h$ and  by $s^i_{t:t+T}\in\mathcal{W}^{T}$ the future trajectory until time $T$. Consider a joint state prediction model $f: \mathcal{W}^{n \times h} \xrightarrow[]{} \mathcal{W}^{n \times T}$, which takes as input the joint states of the agents $\boldsymbol{s}^{1:n}_{t-h:t}$ and predicts the future states $\boldsymbol{\hat{s}}^{1:n}$.

\begin{equation*}
    f \left(s^1_{t-h:t}, \dots s^n_{t-h:t}\right) = (\hat{s}^1_{t:t+T}, \dots \hat{s}^n_{t:t+T}) = \boldsymbol{\hat{s}}^{1:n} 
\end{equation*}


We denote the distribution of future states for an agent $i\in\mathcal{N}$ as $p(\hat{s}_{t:t+T}^{i})$, and the joint distribution of states is represented as $p(\boldsymbol{\hat{s}}^{1:n})$. The prediction model $f: \mathcal{W}^{t \times n} \times \mathcal{W}^{T \times n} \xrightarrow[]{} [0, 1]$ is a conditional distribution; denoting the distribution of the future trajectories given past trajectories of all the agents i.e $f$ corresponds to $p(\boldsymbol{\hat{s}}^{1:n} | s_{t-h:t}^{1:n})$.

\begin{figure*}
    \centering
         \includegraphics[width = 0.98\linewidth]{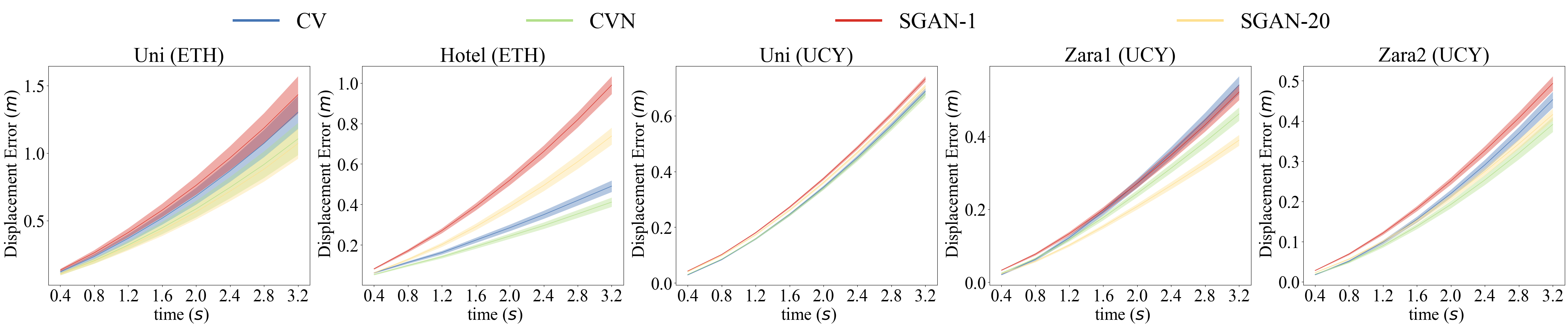}
\caption{Error in trajectory prediction of humans on the ETH~\citep{PellegriniESG09} and UCY~\citep{UCY} datasets. Baselines are referred to from~\citep{scoller2019cv}. Error bars indicate $95\%$ confidence intervals, and the line represents the minimum displacement error across the samples. \label{fig:offline-errors}}
\end{figure*}

\subsection{Probabilistic Trajectory Prediction using S-GAN}\label{sec:predictionsgans}

In this paper, we implement a probabilistic trajectory prediction mechanism $f$ using Social GAN (S-GAN), a state-of-the-art model from~\citet{gupta2018social}. A GAN consists of two neural networks: a generator $\mathcal{G}$ that estimates the data distribution and a discriminator $\mathcal{D}$ that classifies examples as real or fake (generated by $\mathcal{G}$). The generator and discriminator are trained via a min-max game:
\begin{equation}
\begin{aligned}
&\min _{\mathcal{G}} \max _{\mathcal{D}} V(\mathcal{G}, \mathcal{D})= \\
&\mathbb{E}_{x \sim p_{\text {data }}(x)}[\log \mathcal{D}(x)]+\mathbb{E}_{z \sim p_{(z)}}[\log (1-\mathcal{D}(\mathcal{G}(z)))]
\end{aligned}
\label{eq:gan}
\end{equation}
Given training data with distribution $p_{train}$, and latent variable $z \sim \mathcal{N}(0,1)$, the generator takes as input $z$ and outputs a sample in the training distribution i.e. $\mathcal{G}(z) \sim p$. This formulation can be extended to conditional distributions such that given a condition $c$ and latent variable $z \sim \mathcal{N}(0,1)$ as input to $\mathcal{G}$, the output is $\mathcal{G}(z,c) \sim p( \cdot | c)$.

S-GAN \citep{gupta2018social} is conditioned on the past states of all the agents, $s_{t-h:t}^{1:n}$. The generator $\mathcal{G}$ comprises an \textbf{Encoder}, i.e., a recurrent network that takes as input $s_{t-h:t}^i, i\in\mathcal{N}$ and generates latent representations; a \textbf{Pooling} module, that takes as input these representations and agents' relative positions, and generates a pooled representation incorporating multiagent interaction; a \textbf{Decoder}, i.e., a recurrent network which takes as input a latent variable $z$, the latent and pooled representations and generates a sample from the future state distribution $p(\hat{s}^i_{t:t+T})\forall i\in\mathrm{N}$. Using $\mathcal{G}$ and $z \sim \mathcal{N}(0, 1)$, given the past states of all humans $s_{t-h:t}^{1:n}$ we generate samples for the future states $\boldsymbol{\hat{s}}^{1:n}$ by approximating the distribution $f(s_{t-h:t}^{1:n}) = \mathcal{G}(s_{t-h:t}^{1:n}, z) \sim p(\boldsymbol{\hat{s}}^{1:n} | s_{t-h:t}^{1:n})$. 
%
In order to model the distribution of trajectories and diversity in samples from the generator, S-GAN adds an auxiliary variety loss, where it generates $k$ predictions and takes the L2 norm from the best prediction as the loss~\citep{haoqiang2017pointset,gupta2018social}. 


\subsection{Offline Prediction Performance}

\citet{scoller2019cv} compared the Average Displacement Error (ADE) and the Final Displacement Error (FDE) of S-GAN-based prediction against CV prediction and CV prediction with added noise (CVN), showing that the latter ones perform comparably across the scenes in the ETH~\citep{PellegriniESG09} and UCY~\citep{UCY} datasets. In~\figref{fig:offline-errors}, we compare their \emph{multistep} prediction performance (i.e., the L2-norm between the predicted position and the ground truth at each timestep of prediction), which is informative for navigation tasks. 

Similarly to the observations of~\citet{scoller2019cv}, we see that S-GAN's performance is mixed. While it exhibits lower error on Zara1, it ties with CVN and CV on ETH-Uni and Zara2 and it is outperformed by them in Hotel, whereas on UCY-Uni all models perform comparably.
It should be noted that the human behavior featured in these datasets mostly consists of linear segments that can be well approximated by CV/CVN whereas the S-GAN models promise a better generalization to more complex, nonlinear behavior. 

\section{MPC with Probabilistic Multiagent Trajectory Prediction}\label{sec:probmodels}

We integrate the prediction models from Sec.~\ref{sec:prediction} into an MPC for navigation in crowds.

\subsection{MPC for Navigation in Crowds}

We employ a discrete MPC formulation for navigation in a multiagent environment:

\begin{equation}
\begin{split}
\boldsymbol{u}^{*} = \arg & \min_{\boldsymbol{u}\in\boldsymbol{\mathcal{U}}} \mathcal{J}(\boldsymbol{s}, \hat{\boldsymbol{s}}, \hat{\boldsymbol{s}}^{1:n})\\
    s.t.\: & s_{t+1} = g(s_t, u_t)\\
           & (\hat{\boldsymbol{s}},\boldsymbol{\hat{s}}^{1:n}) = f(s^{1:n}_{t-h:t}, s_{t-h:t}),\: i\in\mathcal{N}
   \label{eq:mpc}
\end{split}\mbox{,}
\end{equation}
where: $\boldsymbol{s}=(s_1,\dots,s_{T})$ is a state rollout, acquired by passing a control trajectory $\boldsymbol{u} = (u_0,\dots,u_{T-1})$ drawn from a space of controls $\mathcal{U}$ through the dynamics $g$; $\boldsymbol{\hat{s}}^i=(\hat{s}_1^i,\dots,\hat{s}_{T}^i)$ is a trajectory prediction for agent $i$, extracted using $f$, which takes as input a state history of $h$ timesteps in the past for all the agents, and $\boldsymbol{\hat{s}}^{1:n}=(\boldsymbol{\hat{s}}^1,\dots,\boldsymbol{\hat{s}}^n)$; $\mathcal{J}$ is a cost expressing considerations of safety, efficiency, and human comfort.

\subsection{MPC with Probabilistic Prediction}\label{sec:mpc-costs}



We use the model from Sec.~\ref{sec:predictionsgans}, to jointly estimate the future states of all agents (including the robot) conditioned on their state histories. We integrate this model of into the MPC framework through the following composite cost: 
\begin{equation}
    \begin{split}
    \mathcal{J}^{exp} (\boldsymbol{s}, \boldsymbol{\hat{s}}^{1:n}) = a_g\mathcal{J}_g(\boldsymbol{s}) +\\ \mathbb{E}\big[a_d\mathcal{J}_d(\boldsymbol{s}, \boldsymbol{\hat{s}}^{1:n}) + &a_p\mathcal{J}_p(\boldsymbol{s}, \boldsymbol{\hat{s}}^{1:n}) + a_c\mathcal{J}_{c}(\boldsymbol{s}, \boldsymbol{\hat{s}}^{1:n})\big]
    \end{split}
    \label{eq:pmpc}
    \mbox{,}
\end{equation}  
where: the expectation is taken over the distribution $(\boldsymbol{\hat{s}}, \boldsymbol{\hat{s}}^{1:n}) \sim p(\boldsymbol{\hat{s}}, \boldsymbol{\hat{s}}^{1:n} | s_{t-h:t}, s^{1:n}_{t-h:t})$; the functions $J_g$, $J_d$ account respectively for \emph{progress to goal}, \emph{respect of users' personal space} (see our prior work~\citep{mavrogiannis2021topologyinformed} for detailed definitions), and \emph{prediction consistency}; $a_d, a_p, a_c$ are weights. 

\textbf{Prediction inconsistency cost}. The minimization of the cost:
\begin{equation}
\begin{split}
\mathcal{J}_{c}(\boldsymbol{s}, \hat{\boldsymbol{s}})=
    \mathop{\mathbb{E}}\left[\lVert \boldsymbol{s} - \boldsymbol{\hat{s}}  \rVert\right]
   \label{eq:sccost}
\end{split}\mbox{,}
\end{equation}
matches in expectation the prediction about the robot motion outputted by the model, $\boldsymbol{\hat{s}}$, given its past interactions with the crowd. Since the prediction model has been trained to jointly predict multiagent interactions, by staying close to predictions about its own motion, the robot can also be more confident about the consistency of its predictions about \emph{users}' motion. In practice, this motivates the robot to avoid maneuvers that could surprise users, forcing them to unexpected reactions that would also be hard to predict using the model. 


Overall, the expected cost $\mathcal{J}^{exp}$ enables the controller to probabilistically reason about the quality of candidate trajectories, incorporating a notion of uncertainty over the future human behavior given the robot's intended behavior. 

\subsection{Simulated Experiments}\label{sec:simulations}

\begin{figure}
    \centering
     \begin{subfigure}{.48\linewidth}
     \centering
         \includegraphics[width = \linewidth]{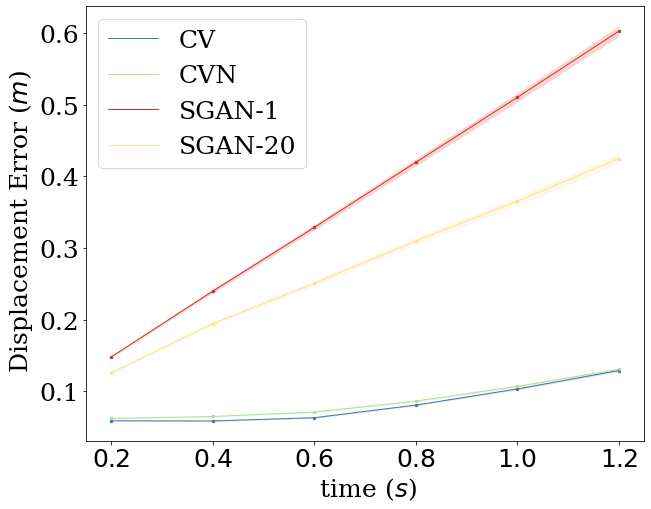}
        \caption{} 
         \label{fig:sim_online-error-plots}
     \end{subfigure}
     \begin{subfigure}{.48\linewidth}
     \centering
         \includegraphics[width = \linewidth]{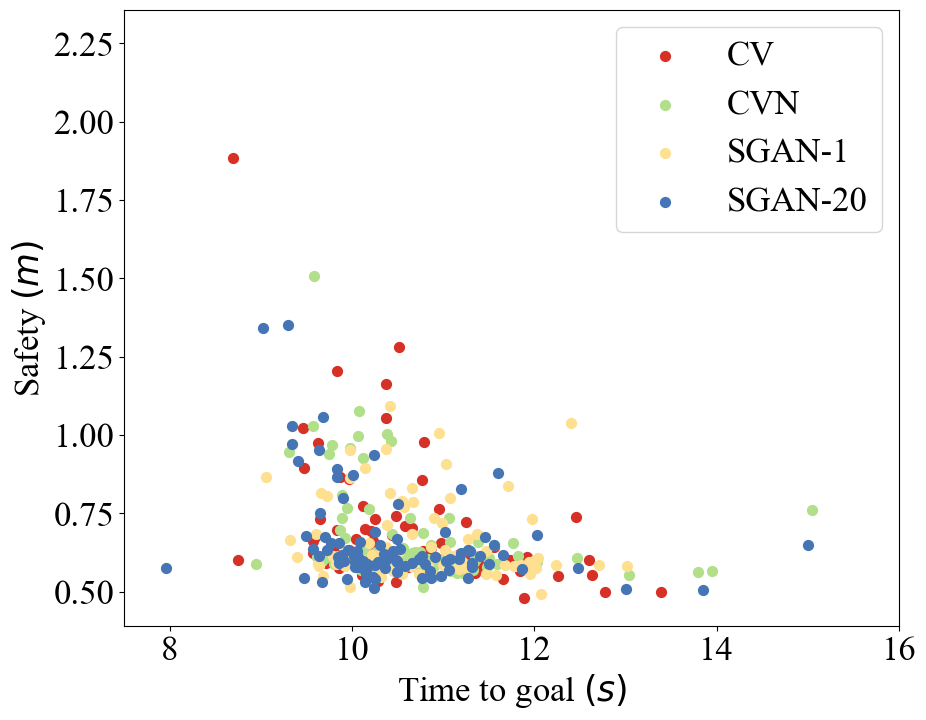}
        \caption{} 
         \label{fig:safety-sim}
     \end{subfigure}
\caption{Simulation results. (\subref{fig:sim_online-error-plots}) Human motion prediction error over time. (\subref{fig:safety-sim}) Safety vs Time to goal. Lines represent minimum displacement errors across the samples and error bands indicate $95\%$ confidence intervals. \label{fig:sim-errors}}
\end{figure}

As a first step towards understanding the impact of prediction accuracy on navigation performance, we instantiated Honda's experimental ballbot~\citep{pathbot,yamane2019,mavrogiannis2021topologyinformed} (see~\figref{fig:introfig}) in a simulated world where human agents were controlled using the ORCA~\citep{ORCA} model. We considered a setting in which three human agents and the robot move across the diagonals of a $3.6\times 4.5 m^2$ workspace (see Table~\ref{tab:resplot}, top left). We evaluated navigation performance in terms of \emph{Safety}, defined as the minimum distance between the robot and human agents (minus the assumed radii of the robot and human agents, both set to $0.3m$) throughout a trial, and \emph{Time to goal}, defined as the time taken by the robot to reach its goal.

\textbf{Algorithms}. We instantiated four different MPC variants, each using a different mechanism for motion prediction:

\emph{MPC with CV prediction}: This baseline approximates the transition function $f$ as a CV model, i.e., $\boldsymbol{s}^i_{t+1} = \boldsymbol{s}^i_{t}+v^i_t\cdot dt$ for agent $i$, where $dt$ represents a timestep; this approximation ignores possible reactions to the motion of other agents.

\emph{MPC with CVN prediction}: This baseline also uses a CV model for prediction but adds noise to the estimate, similarly to the model tested by~\citet{scoller2019cv}.

\emph{MPC with S-GAN-1 prediction}: This baseline uses a single-sample estimate extracted from the S-GAN~\citep{gupta2018social} model. We used the best performing model trained on the ETH dataset.

\emph{MPC with S-GAN-20 prediction}: This baseline uses a 20-sample estimate extracted from the same S-GAN model.

\textbf{Implementation}. We follow an MPC implementation similar to~\citet{brito-ral21}, using a set ${\mathcal{U}}$ of robot control trajectories extracted by propagating the robot with constant velocity towards $10$ subgoals, placed around the robot at fixed orientation intervals of $\frac{\pi}{5}$ and distance of $10m$ for 10 timesteps of size $0.1s$. This parametrization enabled timely response to the dynamic environment: our control loop closed with a frequency of $10Hz$. We tuned all MPC variants through parameter sweeps balancing \emph{Safety} and \emph{Time}.

\textbf{Results}. \figref{fig:sim_online-error-plots} depicts multistep displacement errors across models. We see that the error of S-GAN models is consistently higher than CV/CVN. We suspect that this is because the behavior of ORCA agents often comprises perfectly linear segments which can be effectively approximated using CV-based models; in contrast, the S-GAN models, trained on real-world datasets are less accurate on ORCA agents. However, we see that the superiority of the CV prediction does not translate to superiority in navigation: a scatter plot for safety vs. time to goal for all trials (\figref{fig:safety-sim}) does not show a clear winner, similarly to conducted pairwise U-tests.

\section{Real-World Experiments}\label{sec:evaluation}

\begin{table}
\caption{Average (ADE) and final (FDE) displacement error ($m$) across real-world experiments.}
\label{tab:online_error}
\centering
\begin{tabular}{@{}ccccccc@{}}
\toprule
                 & \multicolumn{2}{c}{Cooperative} & \multicolumn{2}{c}{Aggressive} & \multicolumn{2}{c}{Distracted} \\ \midrule
Prediction Model & ADE            & FDE            & ADE            & FDE           & ADE            & FDE           \\ \midrule
S-GAN-20         &                \textbf{0.257} & \textbf{0.388}                &   \textbf{0.276}             &  \textbf{0.423}             &    \textbf{0.345}            &   \textbf{0.531}            \\
S-GAN-1          &                0.357          & 0.575                &   0.393             &  0.637             &   0.504             &      0.827         \\
CV               &                0.389          & 0.644                &   0.334            &     0.539          &    0.462            &     0.753          \\ \bottomrule
\end{tabular}
\end{table}



As discussed in Sec.~\ref{sec:intro}, benchmarking in a simulated environment --while a widely adopted practice in crowd navigation research-- comes with limitations~\citep{core-challenges2021,fraichard2020}. In this section, we investigate the relationship between prediction and navigation under realistic settings in the lab.

\begin{table*}
\caption{Real-world experiments. Each row shows a different experimental condition: an illustration of the crowd behavior under each condition is shown on the left (users and their goals are shown in blue, whereas the robot and its goal are shown in black color); the multistep prediction error across trials is shown in the middle (error bands indicate $95\%$ confidence intervals); a scatter plot of \emph{Safety} against \emph{Time to goal} is shown on the right.
\label{tab:resplot}}

\centering
\resizebox{\linewidth}{!}{%
\begin{tabular}{@{}ccc@{}}
\toprule
Condition & Prediction Error & \emph{Safety} vs \emph{Time to goal} \\ \midrule

{\label{fig:agents_coop}}{{\includegraphics[height=0.28\linewidth]{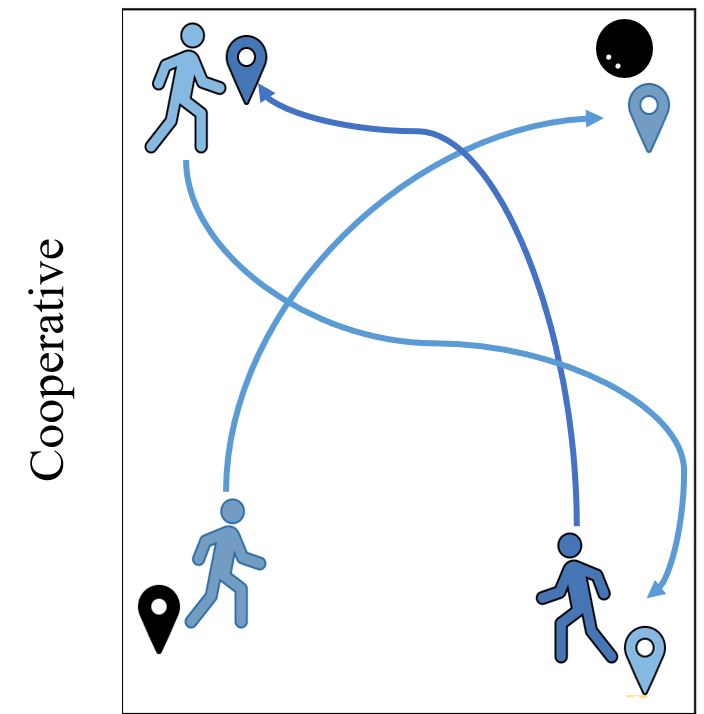}}}     &
{\label{fig:real_online-error-plots}}{\includegraphics[height=0.28\linewidth]{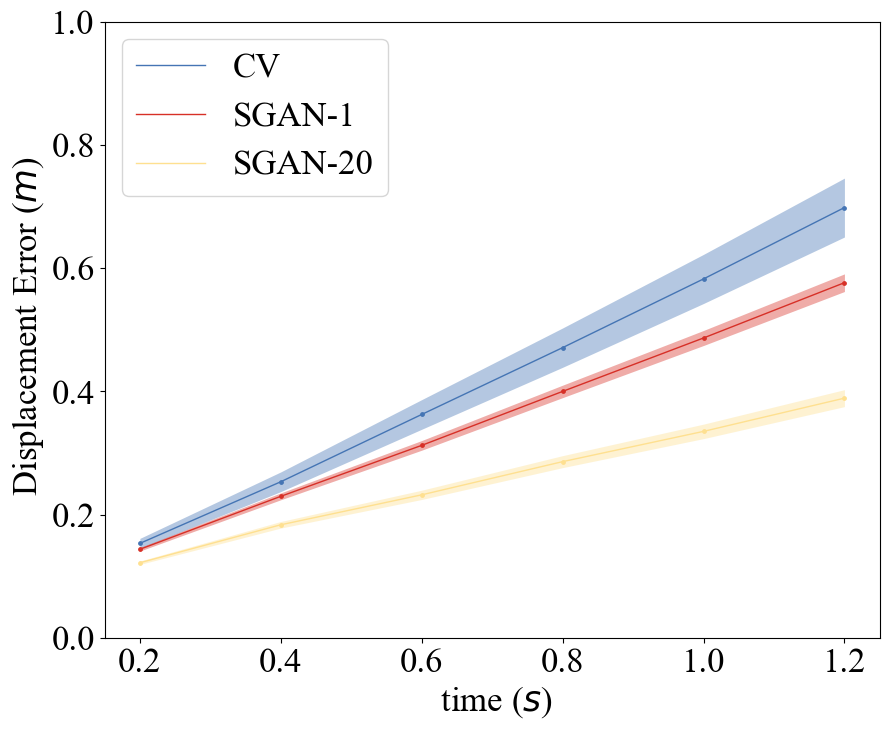}}      &
{\label{fig:eff_coop}}{\includegraphics[height=0.28\linewidth]{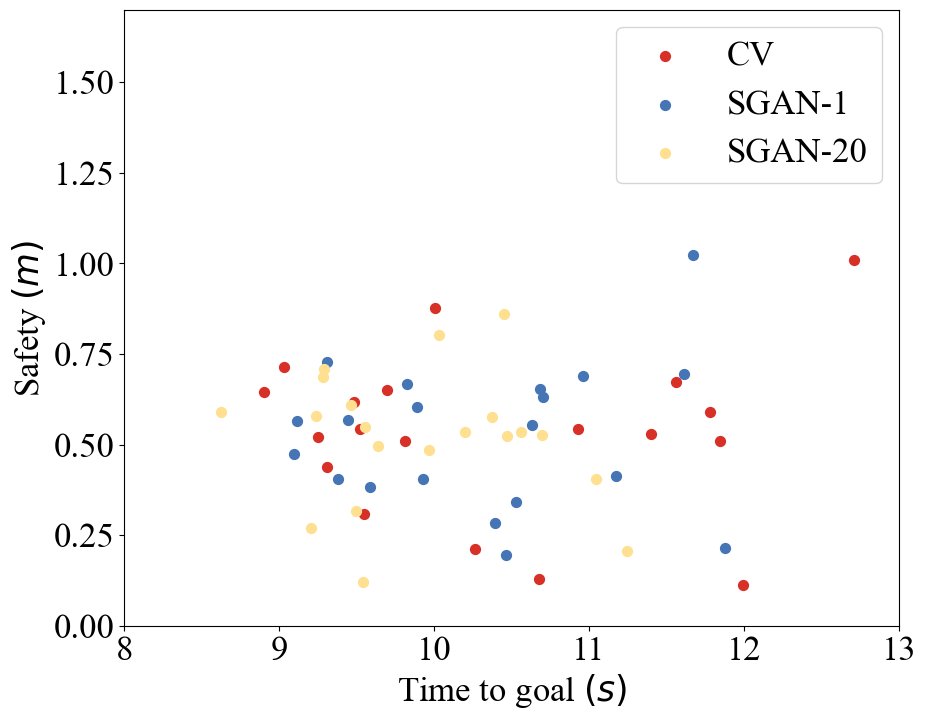}}            \\

{\label{fig:agents_noncoop}}{{\includegraphics[height=0.28\linewidth]{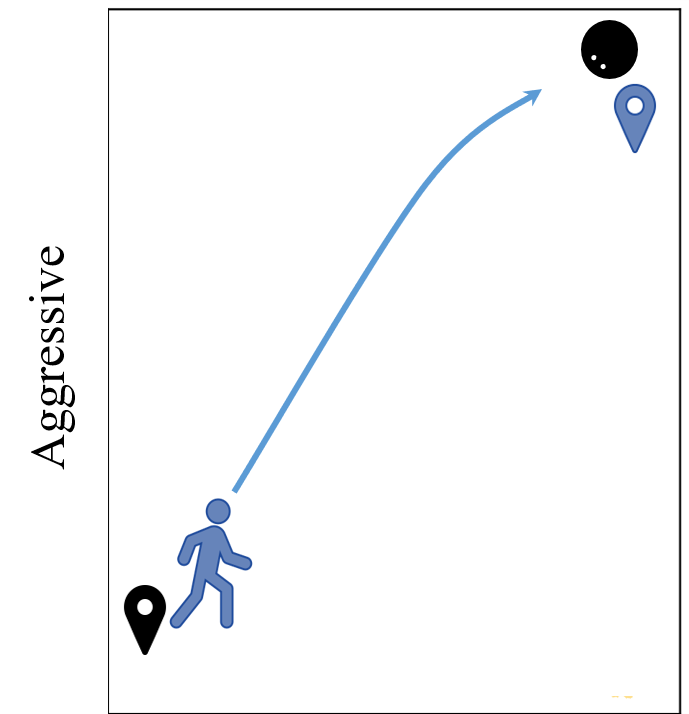}}}   &
{\label{fig:noncoop_error}}{\includegraphics[height=0.28\linewidth]{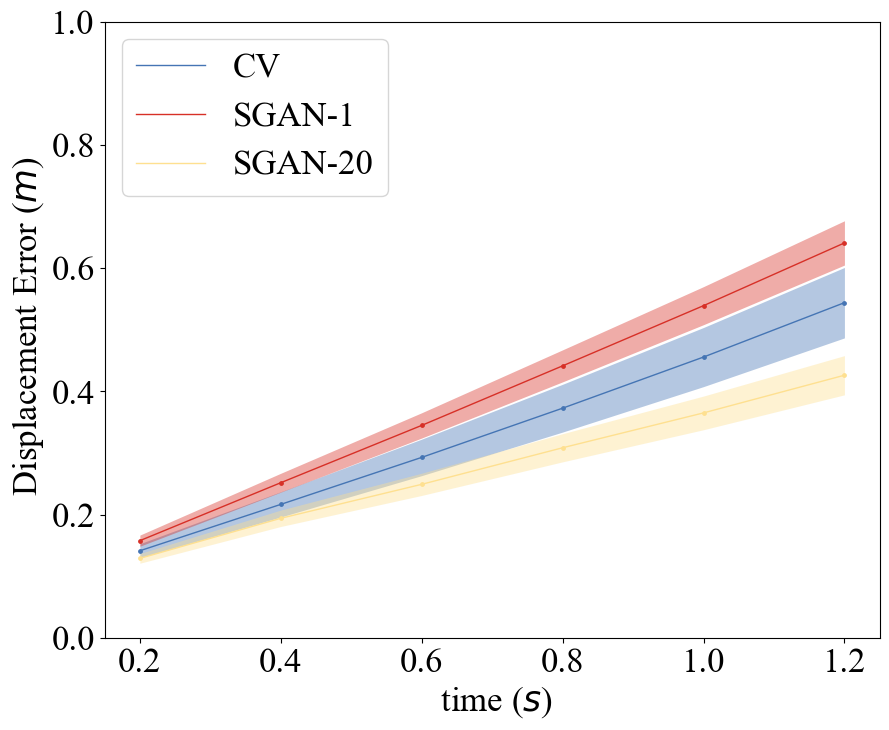}}      &
{\label{fig:eff_noncoop}}{\includegraphics[height=0.28\linewidth]{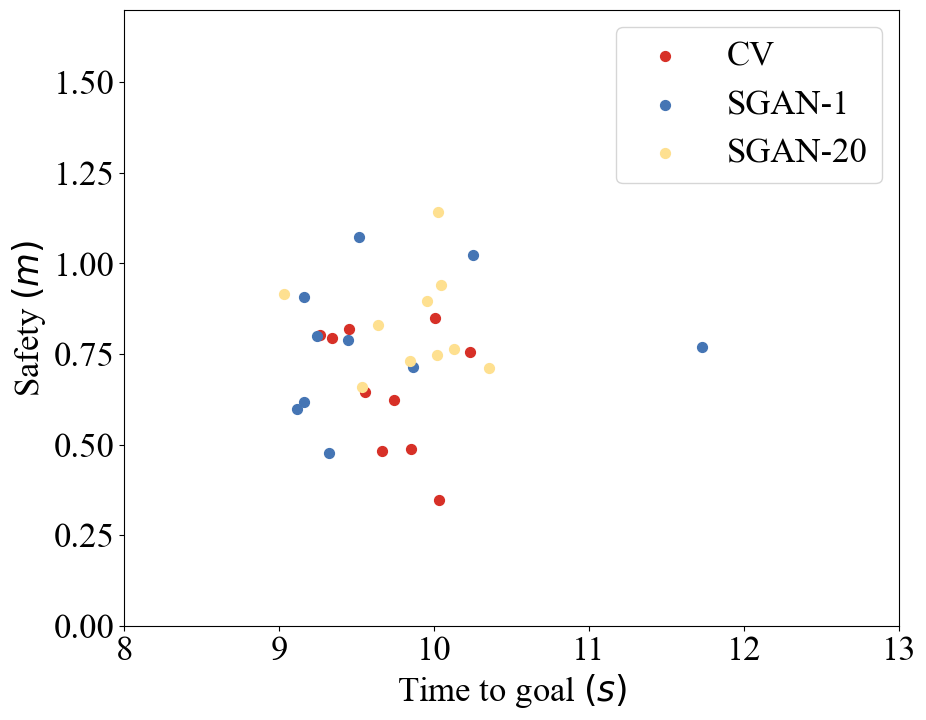}}          \\

{\label{fig:agents_dis}}{{\includegraphics[height=0.28\linewidth]{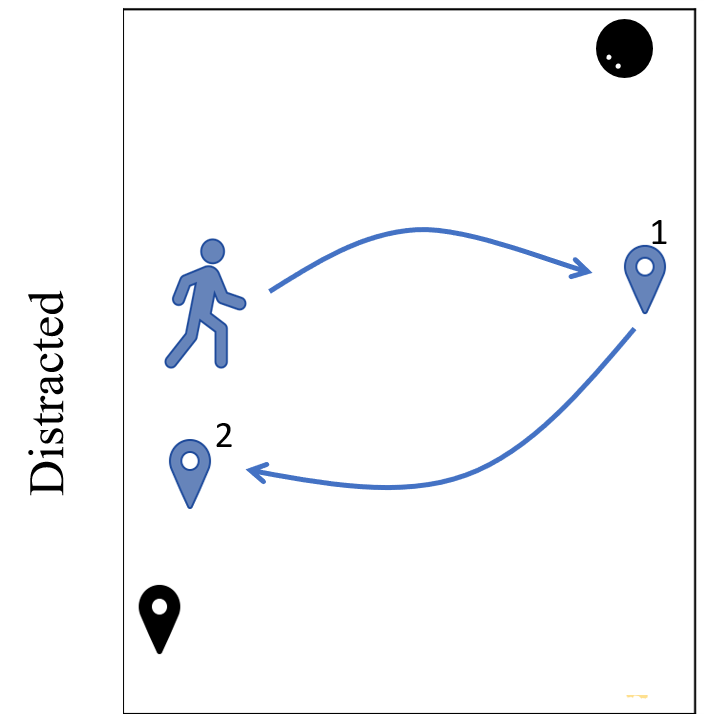}}}    &
{\label{fig:dis_error}}{\includegraphics[height=0.28\linewidth]{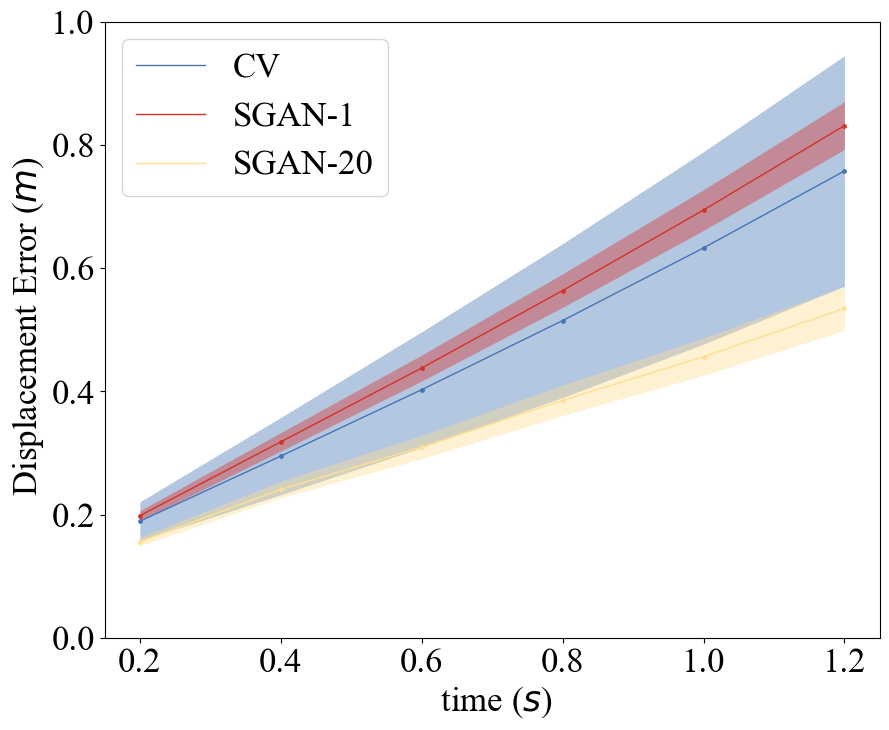}}      &
{\label{fig:eff_dis}}{\includegraphics[height=0.28\linewidth]{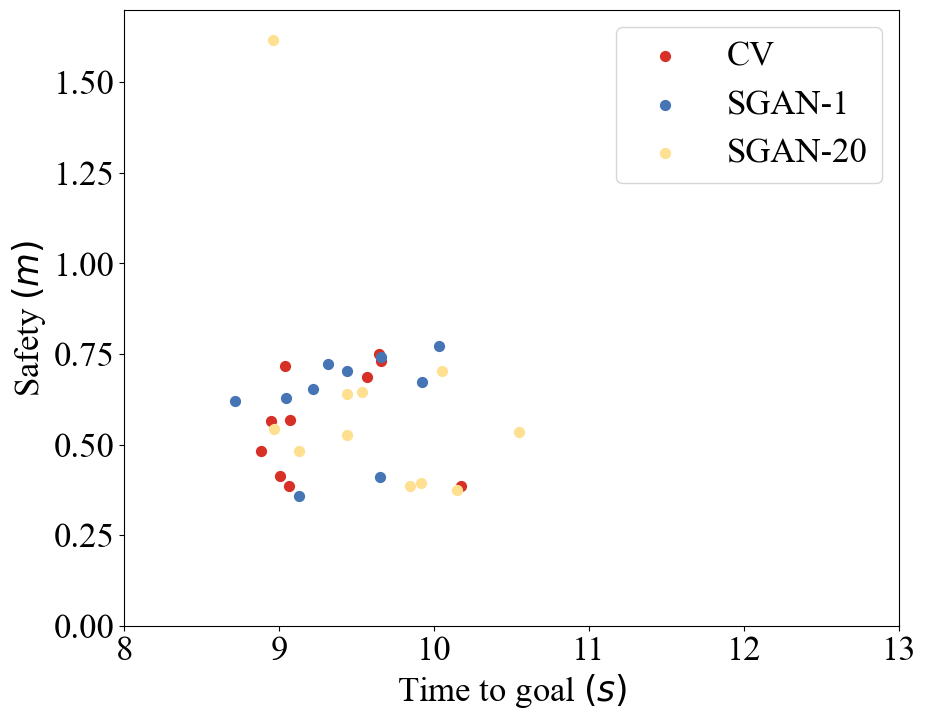}}           \\ 

\bottomrule
\end{tabular}
}
\label{tab:conditions-plots}
\end{table*}

\subsection{Experimental Setup}

We used Honda's experimental ballbot~\citep{pathbot,yamane2019,mavrogiannis2021topologyinformed} (see~\figref{fig:introfig}), and deployed it into a rectangular workspace of area $3.6\times 4.5 m^2$, mirroring out simulation setup.

\textbf{Conditions}. We designed three experimental conditions (shown on the left column of Table~\ref{tab:conditions-plots}) involving robot navigation under different crowd behaviors that a robot could encounter in a crowded space:



\emph{Cooperative}: Three users and the robot move between the corners of the workspace. Users were instructed to navigate naturally with a normal walking speed. 

\emph{Aggressive}: One user and the robot exchange corners. The user was instructed to move straight with normal walking speed without accounting for collisions, i.e., forcing the robot to assume complete responsibility for collision avoidance. 




\emph{Distracted}: One user, starting from the left side, first navigates to the right but quickly moves back to their initial configuration.

Across all conditions, the robot moves between the start and goal points, fixed at $(0,0)$ and $(3.6, 4.5)$ respectively. The preferred speed for the robot is set to $0.8m/s$ which was empirically observed to be a natural speed for users during pilot trials. We conducted real-world experiments under all conditions (20 trials per algorithm for the cooperative condition and 10 trials per algorithm for the rest). 




\textbf{Algorithms}. Across conditions, we compared the performance of the same MPC architecture under three different motion prediction models: CV, S-GAN-1, and S-GAN-20. We did not instantiate a baseline based on CVN since it was shown to perform comparably with CV in simulation.



\textbf{Hypotheses}. While S-GAN models performed worse than CV in a simulated world, their prediction accuracy on real-world datasets~\citep{gupta2018social} (\figref{fig:offline-errors}) appeared promising for operation in the real world. Thus, we expected to see S-GAN models outperform baselines and enable improved navigation performance. We formalized these expectations into the following hypotheses:


\begin{itemize}

    \item [\textbf{H1}:] S-GAN-based prediction is more accurate than CV prediction across all conditions.

    \item[\textbf{H2}:] S-GAN-based prediction enables the MPC to achieve higher navigation performance across all conditions.

    \item[\textbf{H3}:] Lower prediction error generally enables the MPC to achieve higher navigation performance.

\end{itemize}









\begin{figure*}
    \centering
    \includegraphics[width = 0.98\linewidth]{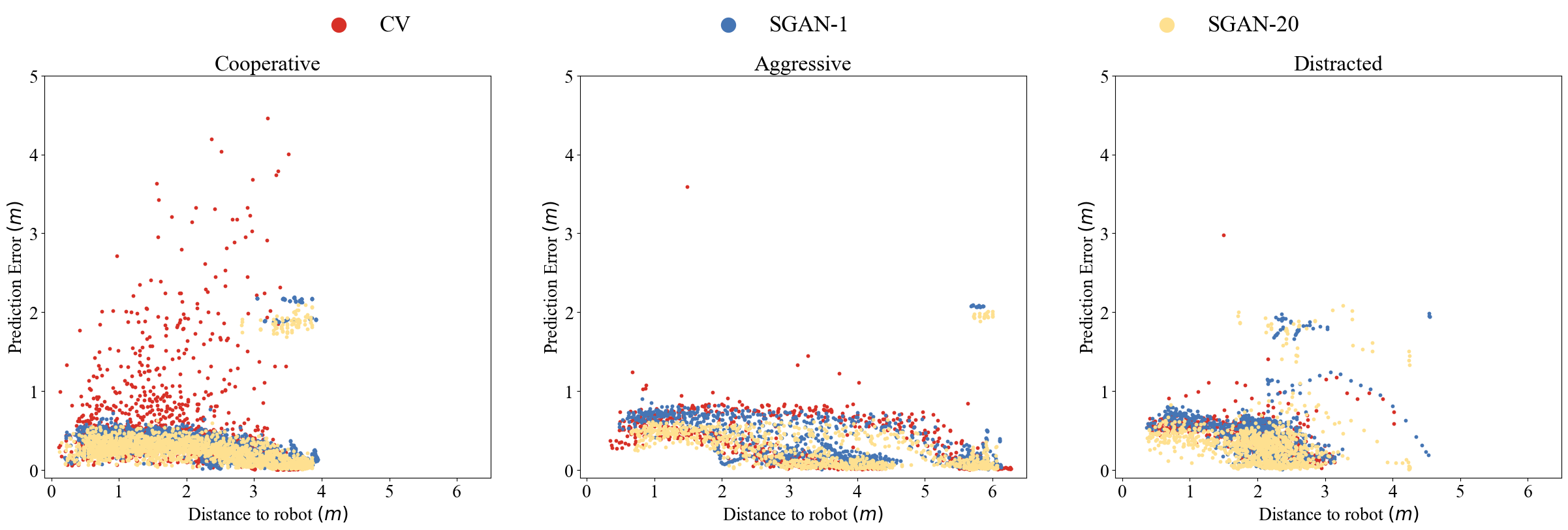}

\caption{Average human trajectory prediction error against distance from human at the time of the prediction.
\label{fig:scatter_distance}}
\end{figure*}

\begin{figure}
    \centering
    \begin{subfigure}{.49\linewidth}
    \centering
        \includegraphics[width=\linewidth]{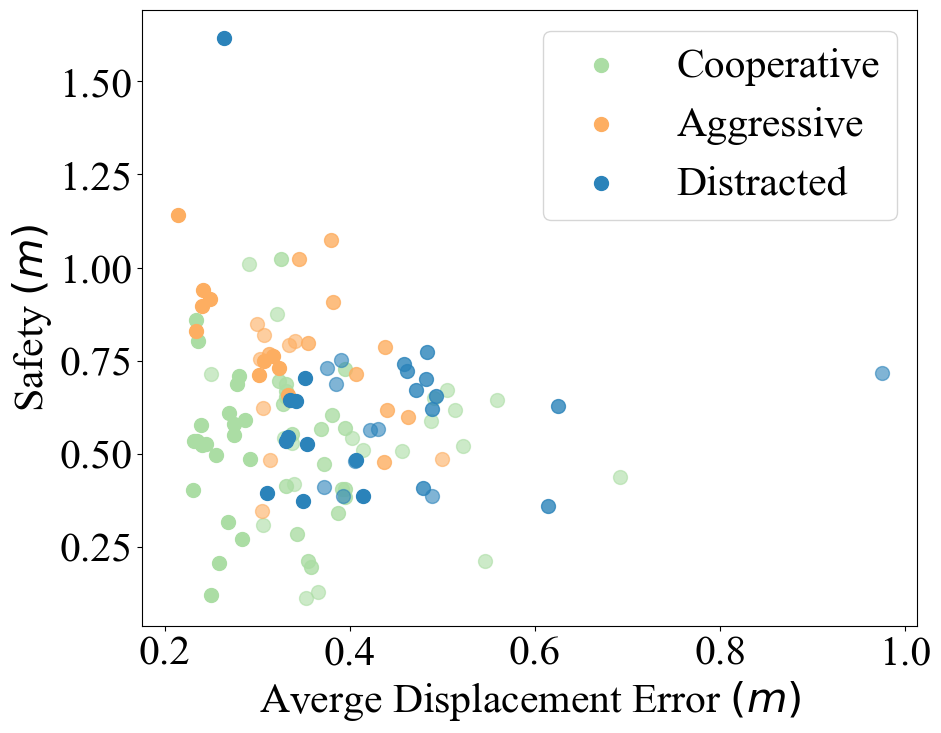}
        \caption{\emph{Safety}\label{fig:scatter-safety}}
    \end{subfigure}
    \begin{subfigure}{.49\linewidth}
    \centering
        \includegraphics[width=\linewidth]{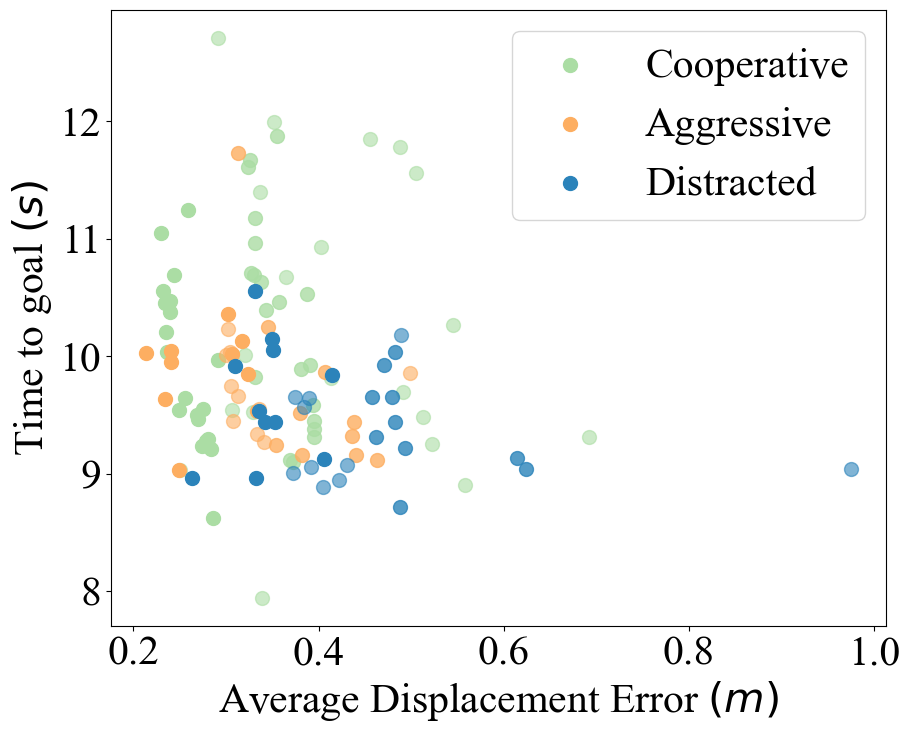}
        \caption{\emph{Time to goal}\label{fig:scatter-efficiency}}
    \end{subfigure}
    \caption{Relationship between prediction performance and navigation performance per trial in the real world.
    \label{fig:scatterplots-H5}}
\end{figure}

\subsection{Results}



\begin{figure*}
    \centering
    \begin{subfigure}{0.3\linewidth}
     \centering
         \includegraphics[width = \linewidth]{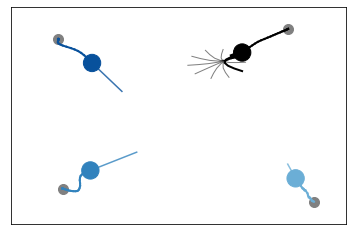}
        \caption{CV.} 
     \end{subfigure}
     \begin{subfigure}{0.3\linewidth}
     \centering
         \includegraphics[width = \linewidth]{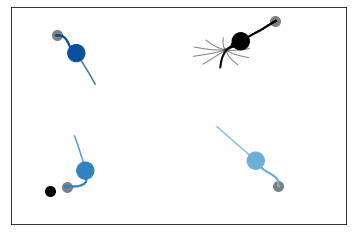}
        \caption{S-GAN-1.} 
     \end{subfigure}
     \begin{subfigure}{0.3\linewidth}
     \centering
         \includegraphics[width = \linewidth]{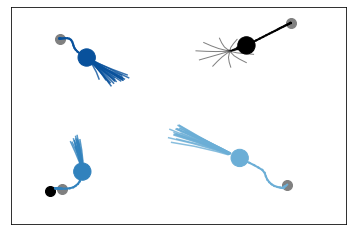}
        \caption{S-GAN-20.} 
     \end{subfigure}

\caption{Instances from lab trials under the cooperative condition for all algorithms. The black dot is the robot and the others are users. Solid lines represent agents' state histories and faded lines are trajectory predictions under each prediction model. The robot's rollouts are shown in gray and the selected one is shown in black.
\label{fig:prediction_samples}}
\end{figure*}

Table~\ref{tab:resplot} shows the multistep prediction error (aggregated values are listed in Table~\ref{tab:online_error})  and the navigation performance distribution per condition. \figref{fig:scatterplots-H5} relates average prediction error per trial to navigation performance. \figref{fig:scatter_distance} connects proximity to the robot to prediction performance. Finally, \figref{fig:prediction_samples} shows how different algorithms make predictions and action decisions in the same scene. Footage from our experiments can be found at \url{https://youtu.be/mzFiXg8KsZ0}.

 
\textbf{H1}. We see that S-GAN-20 outperforms CV and S-GAN-1 in terms of ADE and FDE across all conditions (Table~\ref{tab:online_error}), and exhibits consistently lower multistep prediction error (Table~\ref{fig:real_online-error-plots}, 2nd column). We also see that S-GAN-1 outperforms CV under the cooperative condition but not under the aggressive and distracted conditions. Thus, we find that H1 holds for a strong S-GAN model like S-GAN-20.


\textbf{H2}. From the right column of Table~\ref{fig:eff_coop}, we see that for the cooperative condition, S-GAN-20 is mostly on the left, corresponding to a good time efficiency, and usually higher than the $0.5m$ \emph{Safety} line whereas the other algorithms are more dispersed all over the graph. Under the Aggressive condition, no major differences are observed in terms of time efficiency; S-GAN-20 is often safer than baselines although not consistently superior. In the distracted condition algorithms seem very close to each other. None of these relationships appeared to be statistically significant (pairwise U-tests). Thus, we find no support that the clear superiority in prediction of S-GAN-20 (H1) translates to superiority in navigation, and therefore H2 is rejected.

\textbf{H3}. \figref{fig:scatterplots-H5} shows scatter plots for \emph{Safety} and \emph{Time to goal} against Displacement Error per trial and condition. Across conditions, we see a pattern connecting lower errors to higher safety and lower time to goal. However, this pattern is not definitive: datapoints are scattered across large regions for both navigation metrics. 
Further, as shown in Table~\ref{tab:resplot}, prediction rankings do not transfer clearly to navigation rankings. Thus, we find no support that lower prediction error correlates with improved navigation and H3 is rejected.

\section{Discussion}\label{sec:discussion}


\textbf{Model Transfer}. The high-quality prediction of S-GAN transferred from offline datasets to online performance onboard the robot: S-GAN-20 was consistently more accurate across conditions in the real world (H1). This finding demonstrated the efficacy of the generative machinery of S-GANs for modeling multiagent interactions in pedestrian domains. However, we saw that S-GANs struggled with out-of-distribution behaviors encountered in the ORCA-simulated trials (Sec.~\ref{sec:simulations}). While ORCA behaviors are less representative of real pedestrians, this observation highlights the level of sensitivity of the model to the modes of interaction found in the training dataset. Explicitly introducing structure into prediction through mathematical representations of multiagent interaction~\citep{roh2020corl,Mangalam2021FromGW,SunM-RSS-21} might contribute towards better transfer across a wider range of behavior.

\textbf{Need for even better prediction}. While S-GAN-20 outperformed its baselines, its superiority was not reflected in navigation performance (H2). The CV model and the less performant S-GAN-1 exhibited similar performance on the same MPC. One reason for this could be that S-GAN-20 --while superior on average (Table~\ref{tab:resplot}, Table~\ref{tab:online_error})--was not substantially better for the needs of the task. As shown in~\figref{fig:prediction_samples}, the error of S-GAN-20 is generally below its baselines but still frequently close to the $1m$ line. This is quite high for navigating in our tight workspace ($3.6\times 4.5 m^2$). Thus, it appears that to perform significantly better in navigation within a dense crowd, we might need prediction models with even lower errors.

\textbf{Robot and crowd motion are entangled}. Across models, we saw that prediction performance did not clearly map to navigation performance (H3). In a space as tight as our lab workspace, robot motion is coupled with crowd motion. We accounted for that using a \emph{joint prediction} model, capturing the close unfolding crowd-robot interactions. However, when the MPC forces the robot to deviate from the model's ego-prediction, the resulting robot motion likely violates the validity of the crowd motion prediction. While the prediction inconsistency cost (see~Sec.~\ref{sec:mpc-costs}) motivated the MPC to stay close to the ego prediction, it is likely that the other costs were conflicting on some occasions, leading to situations outside of the model's confidence. An exciting direction for future work involves incorporating explicit formalisms of prediction model confidence into online decision-making.


\textbf{Beyond the Safety-Efficiency tradeoff}. After the lab experiments, users informally stated that MPC with S-GAN was predictable, safer, and more comfortable, but these values are not reflected in the evaluation metrics. While Safety and Efficiency are extensively used as evaluation metrics in social navigation~\citep{core-challenges2021}, they fail to capture important aspects of interaction like smoothness~\citep{social-momentum-thri} or human impressions~\citep{walker2021corl}. This motivates future work on the design of validated metrics capturing critical aspects of interaction during navigation.

\balance
\bibliographystyle{abbrvnat}
\bibliography{references}  

\begin{thebibliography}{44}
\providecommand{\natexlab}[1]{#1}
\providecommand{\url}[1]{\texttt{#1}}
\expandafter\ifx\csname urlstyle\endcsname\relax
  \providecommand{\doi}[1]{doi: #1}\else
  \providecommand{\doi}{doi: \begingroup \urlstyle{rm}\Url}\fi

\bibitem[Becker et~al.(2018)Becker, Hug, H{\"u}bner, and
  Arens]{Becker2018REDAS}
S.~Becker, R.~Hug, W.~H{\"u}bner, and M.~Arens.
\newblock Red: A simple but effective baseline predictor for the {T}raj{N}et
  benchmark.
\newblock In \emph{Workshops of the European Conference on Computer Vision
  (ECCV)}, 2018.

\bibitem[Bhattacharyya et~al.(2019)Bhattacharyya, Hanselmann, Fritz, Schiele,
  and Straehle]{bhattacharya2019confitional}
A.~Bhattacharyya, M.~Hanselmann, M.~Fritz, B.~Schiele, and C.-N. Straehle.
\newblock Conditional flow variational autoencoders for structured sequence
  prediction.
\newblock In \emph{Proceedings of the IEEE/CVF Conference on Computer Vision
  and Pattern Recognition (CVPR)}, 2019.

\bibitem[Biswas et~al.(2022)Biswas, Wang, Silvera, Steinfeld, and
  Admoni]{socnavbench}
A.~Biswas, A.~Wang, G.~Silvera, A.~Steinfeld, and H.~Admoni.
\newblock Soc{N}av{B}ench: A grounded simulation testing framework for
  evaluating social navigation.
\newblock \emph{Transactions on Human-Robot Interaction}, 11\penalty0 (3),
  2022.

\bibitem[Brito et~al.(2021)Brito, Everett, How, and Alonso-Mora]{brito-ral21}
B.~Brito, M.~Everett, J.~P. How, and J.~Alonso-Mora.
\newblock Where to go next: Learning a subgoal recommendation policy for
  navigation in dynamic environments.
\newblock \emph{IEEE Robotics and Automation Letters}, 6\penalty0 (3):\penalty0
  4616--4623, 2021.

\bibitem[Casas et~al.(2018)Casas, Luo, and Urtasun]{CasasLU18}
S.~Casas, W.~Luo, and R.~Urtasun.
\newblock Intentnet: Learning to predict intention from raw sensor data.
\newblock In \emph{Proceedings of the Conference on Robot Learning (CoRL)},
  2018.

\bibitem[Chen et~al.(2019)Chen, Liu, Kreiss, and Alahi]{chen2019crowd}
C.~Chen, Y.~Liu, S.~Kreiss, and A.~Alahi.
\newblock Crowd-robot interaction: Crowd-aware robot navigation with
  attention-based deep reinforcement learning.
\newblock In \emph{Proceedings of the IEEE International Conference on Robotics
  and Automation (ICRA)}, pages 6015--6022, 2019.

\bibitem[Chen et~al.(2020)Chen, Hu, Nikdel, Mori, and
  Savva]{chen2020relational}
C.~Chen, S.~Hu, P.~Nikdel, G.~Mori, and M.~Savva.
\newblock Relational graph learning for crowd navigation.
\newblock In \emph{Proceesings of the IEEE/RSJ International Conference on
  Intelligent Robots and Systems (IROS)}, pages 10007--10013, 2020.

\bibitem[Djuric et~al.(2018)Djuric, Radosavljevic, Cui, Nguyen, Chou, Lin, and
  Schneider]{djuric2018motion}
N.~Djuric, V.~Radosavljevic, H.~Cui, T.~Nguyen, F.-C. Chou, T.-H. Lin, and
  J.~G. Schneider.
\newblock Motion prediction of traffic actors for autonomous driving using deep
  convolutional networks.
\newblock In \emph{Proceedings of the IEEE Winter Conference on Applications of
  Computer Vision (WACV)}, 2018.

\bibitem[Everett et~al.(2018)Everett, Chen, and How]{Everett18_IROS}
M.~Everett, Y.~F. Chen, and J.~P. How.
\newblock Motion planning among dynamic, decision-making agents with deep
  reinforcement learning.
\newblock In \emph{Proceedings of the IEEE/RSJ International Conference on
  Intelligent Robots and Systems (IROS)}, pages 3052--3059, 2018.

\bibitem[Fan et~al.(2017)Fan, Su, and Guibas]{haoqiang2017pointset}
H.~Fan, H.~Su, and L.~Guibas.
\newblock A point set generation network for 3d object reconstruction from a
  single image.
\newblock In \emph{2017 IEEE Conference on Computer Vision and Pattern
  Recognition (CVPR)}, pages 2463--2471, 2017.

\bibitem[Fraichard and Levesy(2020)]{fraichard2020}
T.~Fraichard and V.~Levesy.
\newblock From crowd simulation to robot navigation in crowds.
\newblock \emph{IEEE Robotics and Automation Letters}, 5\penalty0 (2):\penalty0
  729--735, 2020.

\bibitem[Grzeskowiak et~al.(2021)Grzeskowiak, Gonon, Dugas, Paez-Granados,
  Chung, Nieto, Siegwart, Billard, Babel, and Pettré]{Grzeskowiak}
F.~Grzeskowiak, D.~Gonon, D.~Dugas, D.~Paez-Granados, J.~J. Chung, J.~Nieto,
  R.~Siegwart, A.~Billard, M.~Babel, and J.~Pettré.
\newblock Crowd against the machine: A simulation-based benchmark tool to
  evaluate and compare robot capabilities to navigate a human crowd.
\newblock In \emph{Proceedings of the IEEE International Conference on Robotics
  and Automation (ICRA)}, pages 3879--3885, 2021.

\bibitem[Gupta et~al.(2018)Gupta, Johnson, Fei-Fei, Savarese, and
  Alahi]{gupta2018social}
A.~Gupta, J.~Johnson, L.~Fei-Fei, S.~Savarese, and A.~Alahi.
\newblock Social {GAN}: Socially acceptable trajectories with generative
  adversarial networks.
\newblock In \emph{Proceedings of the IEEE/CVF Conference on Computer Vision
  and Pattern Recognition (CVPR)}, pages 2255--2264, 2018.

\bibitem[Helbing and Moln\'ar(1995)]{Helbing95}
D.~Helbing and P.~Moln\'ar.
\newblock Social force model for pedestrian dynamics.
\newblock \emph{Physical Review E}, 51\penalty0 (5):\penalty0 4282--4286, 1995.

\bibitem[Hoermann et~al.(2018)Hoermann, Bach, and
  Dietmayer]{hoermann2017dynamic}
S.~Hoermann, M.~Bach, and K.~Dietmayer.
\newblock Dynamic occupancy grid prediction for urban autonomous driving: A
  deep learning approach with fully automatic labeling.
\newblock In \emph{Proceedings of the International Conference on Robotics and
  Automation (ICRA)}, page 2056–2063, 2018.

\bibitem[Honda(2019)]{pathbot}
Honda.
\newblock Honda {P.A.T.H.} {B}ot, 2019.
\newblock URL \url{https://global.honda/innovation/CES/2019/path_bot.html}.

\bibitem[Kretzschmar et~al.(2016)Kretzschmar, Spies, Sprunk, and
  Burgard]{kretzschmar_ijrr16}
H.~Kretzschmar, M.~Spies, C.~Sprunk, and W.~Burgard.
\newblock Socially compliant mobile robot navigation via inverse reinforcement
  learning.
\newblock \emph{The International Journal of Robotics Research}, 35\penalty0
  (11):\penalty0 1289--1307, 2016.

\bibitem[Lerner et~al.(2007)Lerner, Chrysanthou, and Lischinski]{UCY}
A.~Lerner, Y.~Chrysanthou, and D.~Lischinski.
\newblock Crowds by example.
\newblock \emph{Computer Graphics Forum}, 26\penalty0 (3):\penalty0 655--664,
  2007.

\bibitem[Liu et~al.(2021{\natexlab{a}})Liu, Chen, Liu, and Shi]{liu2021gcn}
C.~Liu, Y.~Chen, M.~Liu, and B.~E. Shi.
\newblock {AVGCN}: Trajectory prediction using graph convolutional networks
  guided by human attention.
\newblock In \emph{Proceedings of the IEEE International Conference on Robotics
  and Automation (ICRA)}, pages 14234--14240, 2021{\natexlab{a}}.

\bibitem[Liu et~al.(2021{\natexlab{b}})Liu, Chang, Liang, Chakraborty, and
  Driggs-Campbell]{liu2020decentralized}
S.~Liu, P.~Chang, W.~Liang, N.~Chakraborty, and K.~Driggs-Campbell.
\newblock Decentralized structural-rnn for robot crowd navigation with deep
  reinforcement learning.
\newblock In \emph{IEEE International Conference on Robotics and Automation
  (ICRA)}, pages 3517--3524, 2021{\natexlab{b}}.

\bibitem[Mangalam et~al.(2021)Mangalam, An, Girase, and
  Malik]{Mangalam2021FromGW}
K.~Mangalam, Y.~An, H.~Girase, and J.~Malik.
\newblock From goals, waypoints \& paths to long term human trajectory
  forecasting.
\newblock \emph{Proceedings of the IEEE/CVF International Conference on
  Computer Vision (ICCV)}, pages 15213--15222, 2021.

\bibitem[Mavrogiannis et~al.(2022)Mavrogiannis, Alves-Oliveira, Thomason, and
  Knepper]{social-momentum-thri}
C.~Mavrogiannis, P.~Alves-Oliveira, W.~Thomason, and R.~A. Knepper.
\newblock Social momentum: Design and evaluation of a framework for socially
  competent robot navigation.
\newblock \emph{Transactions on Human-Robot Interaction}, 11\penalty0 (2),
  2022.

\bibitem[Mavrogiannis et~al.(2023)Mavrogiannis, Balasubramanian, Poddar,
  Gandra, and Srinivasa]{mavrogiannis2021topologyinformed}
C.~Mavrogiannis, K.~Balasubramanian, S.~Poddar, A.~Gandra, and S.~S. Srinivasa.
\newblock Winding through: Crowd navigation via topological invariance.
\newblock \emph{IEEE Robotics and Automation Letters}, 8\penalty0 (1):\penalty0
  121--128, 2023.

\bibitem[{Mavrogiannis} et~al.(2023){Mavrogiannis}, {Baldini}, {Wang}, {Zhao},
  {Trautman}, {Steinfeld}, and {Oh}]{core-challenges2021}
C.~{Mavrogiannis}, F.~{Baldini}, A.~{Wang}, D.~{Zhao}, P.~{Trautman},
  A.~{Steinfeld}, and J.~{Oh}.
\newblock {Core Challenges of Social Robot Navigation: A Survey}.
\newblock \emph{Transactions on Human-Robot Interaction}, 2023.

\bibitem[Nikhil and Morris(2018)]{Nikhil2018ConvolutionalNN}
N.~Nikhil and B.~T. Morris.
\newblock Convolutional neural network for trajectory prediction.
\newblock In \emph{Workshops of the European Conference on Computer Vision
  (ECCV)}, 2018.

\bibitem[Park et~al.(2016)Park, Hwang, Niu, and Shi]{soo2016egocentric}
H.~S. Park, J.-J. Hwang, Y.~Niu, and J.~Shi.
\newblock Egocentric future localization.
\newblock In \emph{Proceedings of the IEEE/CVF Conference on Computer Vision
  and Pattern Recognition (CVPR)}, pages 4697--4705, 2016.

\bibitem[Pellegrini et~al.(2009)Pellegrini, Ess, Schindler, and
  Van~Gool]{PellegriniESG09}
S.~Pellegrini, A.~Ess, K.~Schindler, and L.~Van~Gool.
\newblock You'll never walk alone: Modeling social behavior for multi-target
  tracking.
\newblock In \emph{Proceedings of the IEEE/CVF International Conference on
  Computer Vision (ICCV)}, pages 261--268, 2009.

\bibitem[{Pirk} et~al.(2022){Pirk}, {Lee}, {Xiao}, {Takayama}, {Francis}, and
  {Toshev}]{pirk2022protocol}
S.~{Pirk}, E.~{Lee}, X.~{Xiao}, L.~{Takayama}, A.~{Francis}, and A.~{Toshev}.
\newblock {A Protocol for Validating Social Navigation Policies}.
\newblock \emph{arXiv e-prints}, 2022.

\bibitem[Roh et~al.(2020)Roh, Mavrogiannis, Madan, Fox, and
  Srinivasa~S.]{roh2020corl}
J.~Roh, C.~Mavrogiannis, R.~Madan, D.~Fox, and S.~Srinivasa~S.
\newblock Multimodal trajectory prediction via topological invariance for
  navigation at uncontrolled intersections.
\newblock In \emph{Proceedings of the Conference on Robot Learning}, 2020.

\bibitem[Rudenko et~al.(2020)Rudenko, Palmieri, Herman, Kitani, Gavrila, and
  Arras]{rudenko2019-predSurvey}
A.~Rudenko, L.~Palmieri, M.~Herman, K.~M. Kitani, D.~M. Gavrila, and K.~O.
  Arras.
\newblock Human motion trajectory prediction: a survey.
\newblock \emph{The International Journal of Robotics Research}, 39\penalty0
  (8):\penalty0 895--935, 2020.

\bibitem[Sadeghian et~al.(2019)Sadeghian, Kosaraju, Sadeghian, Hirose,
  Rezatofighi, and Savarese]{sophie19}
A.~Sadeghian, V.~Kosaraju, A.~Sadeghian, N.~Hirose, H.~Rezatofighi, and
  S.~Savarese.
\newblock Sophie: An attentive gan for predicting paths compliant to social and
  physical constraints.
\newblock In \emph{Proceedings of the IEEE/CVF Conference on Computer Vision
  and Pattern Recognition (CVPR)}, 2019.

\bibitem[Salzmann et~al.(2020)Salzmann, Ivanovic, Chakravarty, and
  Pavone]{trajectron}
T.~Salzmann, B.~Ivanovic, P.~Chakravarty, and M.~Pavone.
\newblock Trajectron++: Dynamically-feasible trajectory forecasting with
  heterogeneous data.
\newblock In \emph{Proceedings of the European Conference on Computer Vision
  (ECCV)}, pages 683--700, 2020.

\bibitem[Sathyamoorthy et~al.(2020)Sathyamoorthy, Liang, Patel, Guan, Chandra,
  and Manocha]{sathyamoorthydensecavoid}
A.~J. Sathyamoorthy, J.~Liang, U.~Patel, T.~Guan, R.~Chandra, and D.~Manocha.
\newblock Dense{CA}void: Real-time navigation in dense crowds using
  anticipatory behaviors.
\newblock \emph{Proceedings of the IEEE International Conference on Robotics
  and Automation (ICRA)}, pages 11345--11352, 2020.

\bibitem[Schöller et~al.(2020)Schöller, Aravantinos, Lay, and
  Knoll]{scoller2019cv}
C.~Schöller, V.~Aravantinos, F.~Lay, and A.~Knoll.
\newblock What the constant velocity model can teach us about pedestrian.
\newblock \emph{IEEE Robotics and Automation Letters}, 5\penalty0 (2):\penalty0
  1696--1703, 2020.

\bibitem[Sun et~al.(2021)Sun, Baldini, Trautman, and Murphey]{SunM-RSS-21}
M.~Sun, F.~Baldini, P.~Trautman, and T.~Murphey.
\newblock {Move Beyond Trajectories: Distribution Space Coupling for Crowd
  Navigation}.
\newblock In \emph{Proceedings of Robotics: Science and Systems}, 2021.

\bibitem[Trautman et~al.(2015)Trautman, Ma, Murray, and Krause]{trautmanijrr}
P.~Trautman, J.~Ma, R.~M. Murray, and A.~Krause.
\newblock Robot navigation in dense human crowds: Statistical models and
  experimental studies of human-robot cooperation.
\newblock \emph{International Journal of Robotics Research}, 34\penalty0
  (3):\penalty0 335--356, 2015.

\bibitem[Tsoi et~al.(2021)Tsoi, Hussein, Fugikawa, Zhao, and
  V\'{a}zquez]{Tsoi_2021_Sean_EP}
N.~Tsoi, M.~Hussein, O.~Fugikawa, J.~D. Zhao, and M.~V\'{a}zquez.
\newblock An approach to deploy interactive robotic simulators on the web for
  {HRI} experiments: Results in social robot navigation.
\newblock In \emph{Proceedings of the IEEE/RSJ International Conference on
  Intelligent Robots and Systems (IROS)}, page 7528–7535, 2021.

\bibitem[van~den Berg et~al.(2011)van~den Berg, Guy, Lin, and Manocha]{ORCA}
J.~van~den Berg, S.~J. Guy, M.~Lin, and D.~Manocha.
\newblock Reciprocal n-body collision avoidance.
\newblock In \emph{Robotics Research}, pages 3--19. Springer Berlin Heidelberg,
  2011.

\bibitem[Walker et~al.(2021)Walker, Mavrogiannis, Srinivasa, and
  Cakmak]{walker2021corl}
N.~Walker, C.~Mavrogiannis, S.~S. Srinivasa, and M.~Cakmak.
\newblock Influencing behavioral attributions to robot motion during task
  execution.
\newblock In \emph{Proceedings of the Conference on Robot Learning (CoRL)},
  2021.

\bibitem[Wang et~al.(2021)Wang, Mavrogiannis, and Steinfeld]{wang2021corl}
A.~Wang, C.~Mavrogiannis, and A.~Steinfeld.
\newblock Group-based motion prediction for navigation in crowded environments.
\newblock In \emph{Proceedings of the Conference on Robot Learning (CoRL)},
  2021.

\bibitem[{Wang} et~al.(2022){Wang}, {Biswas}, {Admoni}, and
  {Steinfeld}]{tbd-dataset}
A.~{Wang}, A.~{Biswas}, H.~{Admoni}, and A.~{Steinfeld}.
\newblock {Towards Rich, Portable, and Large-Scale Pedestrian Data Collection}.
\newblock \emph{arXiv e-prints}, Mar. 2022.

\bibitem[Yamane and Kurosu(2020)]{yamane2019}
K.~Yamane and C.~Kurosu.
\newblock Stable balance controller, March 2020.
\newblock US Patent 16/375,111.

\bibitem[Zhang et~al.(2019)Zhang, Ouyang, Zhang, Xue, and Zheng]{zhang19srlstm}
P.~Zhang, W.~Ouyang, P.~Zhang, J.~Xue, and N.~Zheng.
\newblock {SR}-{LSTM}: State refinement for {LSTM} towards pedestrian
  trajectory prediction.
\newblock In \emph{Proceedings of the IEEE/CVF Conference on Computer Vision
  and Pattern Recognition (CVPR)}, pages 12077--12086, 2019.

\bibitem[Ziebart et~al.(2009)Ziebart, Ratliff, Gallagher, Mertz, Peterson,
  Bagnell, Hebert, Dey, and Srinivasa]{ziebart2009}
B.~D. Ziebart, N.~Ratliff, G.~Gallagher, C.~Mertz, K.~Peterson, J.~A. Bagnell,
  M.~Hebert, A.~K. Dey, and S.~Srinivasa.
\newblock Planning-based prediction for pedestrians.
\newblock In \emph{Proceedings of the IEEE/RSJ International Conference on
  Intelligent Robots and Systems (IROS)}, pages 3931--3936, 2009.

\end{thebibliography}

\end{document}